# Modus Tollens and Counterfactuals and Counterfactual Reasoning Based on Three Types of Negation

Zhenghua Pan

School of Science, Institute of Mathematics and Physics, Jiangnan University, Wuxi 214122, China

**Abstract:** Modus Tollens (MT) is a classical logical inference rule, while counterfactuals are hypothetical statements that are contrary to facts, and counterfactual reasoning is a process of reasoning based on counterfactuals. Negation is an indispensable core concept in them. In this paper, based on the logical systems LCOI&PLCOI with contradictory negation, opposite negation and intermediary negation, we propose three variants of Modus Tollens corresponding to distinct negation types, namely "$MT_C$: Modus Tollens based on contradictory negation", "$MT_O$: Modus Tollens based on opposite negation", and "$MT_I$: Modus Tollens based on intermediary negation". We define the implications within $MT_C$, $MT_O$ and $MT_I$, provide the truth value algorithms of $MT_C$, $MT_O$ and $MT_I$, and discuss the reducibility of these algorithms. To incorporate these three types of negation into counterfactuals and counterfactual reasoning, we differentiate counterfactuals into two types based on whether they possess logical negation, thereby proposing three counterfactuals and counterfactuals reasoning based on different logical negations.

In this paper, we further argue that the three counterfactuals reasoning based on different logical negations have the same inference form as $MT_C$, $MT_O$ and $MT_I$, respectively. In other words, they share the same inference structure. As a result, the truth value algorithms for $MT_C$, $MT_O$ and $MT_I$ can be as the truth value algorithms for the three counterfactuals reasoning based on different logical negations. The algorithms indicates that if the first premise of the reasoning is true, the truth values of the reasoning conclusions are identical to the truth values of the three negative premises in the reasoning premises, respectively. This reflects the consistency and accuracy of the truth value algorithms.

This work was supported by the National Natural Science Foundation of China [60575038, 60973156, 61375004]; the State Key Laboratory for Novel Software Technology, Nanjing University, P.R. China [KFKT2020B01]

## 1 Introduction

"Negation" is a common feature of natural language and an important fundamental concept in knowledge [1]. Broadly speaking, negation connects an expression E with another expression whose meaning is, in some sense, opposed to that of E [2]. Therefore, the key challenge in understanding negation is to identify the meaning that is in some way opposed to E—this is a semantically complex and highly vague task [3]. Due to its complexity, the various linguistic forms it can take, and the different ways it interacts with words within its scope, the computational processing of negation remains unresolved and has become a research hotspot across different knowledge domains [4].

Modus Tollens (MT) is a classical rule of logical inference, also known as the method of denying the consequent. Its basic structure: If P then Q, not Q, therefore not P. This form of inference has important applications in fields such as logic, mathematics, philosophy, computer science, and artificial intelligence. The core idea of MT is to derive the negation of the premise (antecedent) of a conditional proposition by denying its result (consequent). It can be said that MT has a direct logical relationship with negation. Negation is not only a key part of MT but also reveals the important mechanism of how conditional reasoning works in reverse [5-9].

Counterfactuals and counterfactual reasoning are cognitive processes of human thought and have long been subjects of extensive research in the fields of philosophy, logic, and artificial intelligence. A counterfactual refers to a hypothetical statement that is contrary to the facts, often used to describe a "what if things had not happened as they actually did" scenario. Counterfactual reasoning is a process that involves reasoning based on counterfactuals, with its basic structure being the negation of facts to construct a hypothetical premise that supports hypothetical reasoning. Negation occupies a central position in both counterfactual statements and reasoning [10-15].

In summary, in both Modus Tollens and counterfactuals with counterfactual reasoning, negation is an indispensable core concept. Modus Tollens employs negation to form valid inferences, while counterfactual reasoning uses conditional negation to explore possible outcomes or inferences. The negation in Modus Tollens typically employs the expression of negation in classical logic. However, in counterfactuals and counterfactual reasoning, the semantic model of negation is not based on a single form of negation logic. The complexity of counterfactual reasoning necessitates considering various forms of negation and different contexts. Non-classical logic, possible world's semantics, conditional logic, causal logic, and modal logic together explain the mechanisms and inference rules of negation in counterfactuals and counterfactual reasoning, forming the foundation for understanding counterfactual reasoning [16-19].

In the development of artificial intelligence, new demands have arisen for the understanding and handling of "negation". Some studies advocate for distinguishing between different types of negation in information processing [20-29]. In these studies, researchers such as Wagner [21] and Analyti et al. [22] argue that the concept of negation plays a special role in any information computational system, where negated information holds equal value to affirmed information. They propose distinguishing between "strong negation" and "weak negation" in information computational systems, with strong negation representing 'explicit falsity' and weak negation denoting 'non-truth'. Kaneiwa argues that description logic should differentiate between two types of negation [26], while Ferré introduces a distinction among "negation", "opposition" and "possibility" in Logical Concept Analysis [27]. Almeida explores lattices with different negation operations and their properties and structures, providing a new perspective for the normative extension of lattices and relational representation [28]. Vinkov et al. propose a stepwise theoretical formalism that includes two types of negation, applicable to a domain of Active Logic [29]. These studies on different types of negation in information share a common feature: they primarily focus on the practical needs of knowledge processing, proposing methods for distinguishing and handling different negations in knowledge from a semantic perspective. However, they do not fundamentally recognize and differentiate the various forms of negation in knowledge, nor do they examine the theoretical foundations that could reflect the properties, relationships, and laws of these different negations [30-34].

In references [33] and [34], we propose that there are three distinct types of negation in knowledge from a conceptual perspective: contradictory negation, opposite negation and intermediary negation, and establish the set SCOI and logic LCOI&PLCOI with three kinds of negation. In this study grounded in the LCOI&PLCOI framework, we propose three variants of Modus Tollens corresponding to distinct negation types, namely "$MT_C$: Modus Tollens based on contradictory negation", "$MT_O$: Modus Tollens based on opposite negation", and "$MT_I$: Modus Tollens based on intermediary negation". We define the implications within $MT_C$, $MT_O$ and $MT_I$, provide the truth value algorithms of $MT_C$, $MT_O$ and $MT_I$, and discuss the reducibility of these algorithms. We introduce contradictory negation, opposite negation and intermediary negation into counterfactual and counterfactual reasoning, presenting three types of counterfactuals and counterfactual reasoning based on different logical negations. We also discussed the relationship between $MT_C$, $MT_O$, $MT_I$ and the three types of counterfactual reasoning based on different logical negations in terms of their inference structure, as well as the truth value algorithms of three counterfactual reasoning based on different logical negations.

The main contributions of this paper are as follows:

1. Proposing three variants of Modus Tollens corresponding to distinct negation types, namely '$MT_C$: Modus Tollens based on contradictory negation', '$MT_O$: Modus Tollens based on opposite negation', and '$MT_I$: Modus Tollens based on intermediary negation', and give their truth value algorithms.
2. Proposing three counterfactuals and counterfactual reasoning based on different logical negations.
3. It has been confirmed that the three types of counterfactual reasoning based on different logical negations have the same inference form as MTC, MTO, and MTI, respectively. That is, they have the same inference structure.
4. The truth value algorithms for the three types of counterfactual reasoning based on different logical negations indicates that if the first premise of the reasoning is true, the truth values of the reasoning conclusions are identical to the truth values of the three negative premises in the reasoning premises. This reflects the consistency and accuracy of the truth value algorithms

The structure of this paper is as follows. Section 2 introduces three types of negation in knowledge and their characteristics. Section 3 presents the basic theory of the logic LCOI&PLCOI with three kinds of negation. In Section 4, we propose three types of Modus Tollens based on different negations, grounded in LCOI&PLCOI. Section 5 introduces the three types of negation into counterfactuals and counterfactual reasoning, and proposes three different counterfactuals and counterfactual reasoning approaches based on distinct logical negations. Additionally, it discusses the algorithms for counterfactual reasoning based on these different logical negations. Section 6 summarizes the main research findings of this paper and outlines future work.

## 2 Three types of negation in knowledge and their characteristics

Knowledge is the mental product of human understanding of the objective world, while concepts are the foundation of thought [35]. From a philosophical perspective, the construction of any knowledge begins with the observation and understanding of things through existing concepts. From the standpoint of artificial intelligence, a fundamental issue in knowledge representation and reasoning is the study and handling of concepts[36]. The research on knowledge, especially common sense and non-standard knowledge, in artificial intelligence involves distinguishing, expressing, judging, and reasoning about concepts and their relationships, which are fundamental issues [37]. In references [33] and [34], we distinguish between "clear concept" and "fuzzy concept" at the conceptual level, fully understanding the "contradictions" and "oppositions" within the concepts, thereby proposing that there are three different types of negation in the concepts: contradictory negation, opposite negation, and intermediary negation. In this section, we briefly outline these three different types of negation and their characteristics.

The following three different forms of negation exist in both clear and fuzzy concepts:

(1) *Contradictory Negation*. For a species concept under a genus concept, another species concept that has a contradictory relationship with it constitutes a form of negation. We refer to this type of negation as "contradictory negation". In this form of negation, the intensions (connotations) of the two species concepts mutually negate each other, the extensions (denotations) are mutually exclusive (either one or the other), and the sum of the extensions equals the extension of the genus concept. From this, it can be known that the negation in classical logic is precisely this kind of negation.

(2) *Opposite Negation*. For a species concept under a genus concept, another species concept that has an oppositional relationship with it constitutes another form of negation. We refer to this type of negation as "opposite negation". In this form of negation, the intensions of the two species concepts mutually negate each other and exhibit the greatest difference in intension, but their extensions are not mutually exclusive (not either-or), and the sum of their extensions is less than the extension of the genus concept.

(3) *Intermediary Negation*. The intermediary concept between opposite concepts constitutes a (weak) form of negation of the opposite concepts. We refer to this type of negation as "intermediary negation". In this form of negation, the opposite concepts transition through the intermediary concept and the sum of their extensions equals the extension of the genus concept.

We must point out that in reality there are cases where "contradiction" and "opposition" appear identical. Such situations should be understood as contradiction rather than opposition. For example, under the genus concept of 'real numbers', the species concepts 'rational numbers' and 'irrational numbers' (i.e., non-rational numbers) are both contradictory and oppositional. However, since there is no "intermediary" between rational and irrational numbers, they are not oppositional concepts.

To fully understand the meaning of the above three kinds of negation, we further discuss their characteristics in terms of both the intension of the concepts as well as their extensional relations.

(1) Contradictory Negation in Clear Concepts (CNC)

Characteristics of CNC: Extensions are clear, either this or that, and the sum of extensions is equal to the extension of the genus concept.

For example, the positive integer and non-positive integer under the genus concept of "integer" are clear concepts, while the non-positive integer is the contradictory negation of positive integer. The diagram illustrating the extensional relationship between them is shown below (Figure 1).

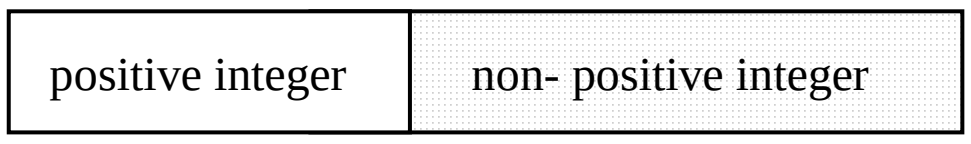

**Fig. 1** The extensional relationship between positive integer and non-positive integer

(2) Opposite Negation in Clear Concepts (ONC)

Characteristics of ONC: Extensions are clear, not "either this or that", and the sum of the extensions is less than the extension of the genus concept.

For example, the positive integer and negative integer under the genus concept of "integer" are clear concepts, while the negative integer is the opposite negation of positive integer. The diagram illustrating the extensional relationship between them is shown below (Figure 2).

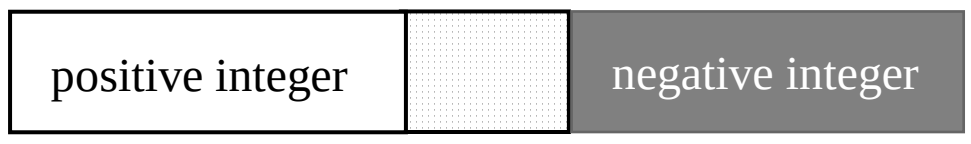

**Fig. 2** The extensional relationship between positive integer and negative integer

(3) Intermediary Negation in Clear Concepts (INC)

Characteristics of INC: Extensions are clear, opposing sides transition to each other through ‘intermediaries’, and the sum of the extensions equals the extension of the genus concept.

For example, zero, positive integer and negative integer are clear concepts under the genus concept of “integer”. Zero serves as an ‘intermediary’ between positive integer and negative integer, and it is the intermediary negation of both positive integer and negative integer. The diagram illustrating the extensional relationship between them is shown below (Figure 3).

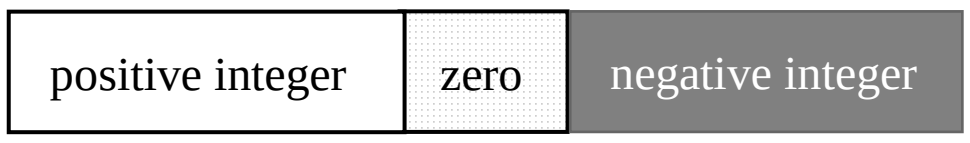

**Fig. 3** The extensional relationship between positive integers, negative integers, and zero.

(4) Contradictory Negation in Fuzzy Concepts (CNF)

Characteristics of CNF: Extensions are not clear, either this or that, and the sum of extensions is equal to the extension of the genus concept.

For example, the daytime and non-daytime under the genus concept of “day” are fuzzy concepts, while the non-daytime is the contradictory negation of daytime. The diagram illustrating the extensional relationship between them is shown below (Figure 4).

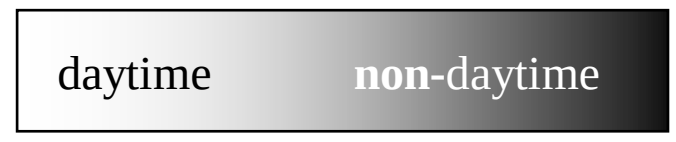

**Fig. 4** The extensional relationship between daytime and non-daytime

(5) Opposite Negation in Fuzzy Concepts (ONF)

Characteristics of ONF: Extensions are not clear, not “either this or that”, and the sum of the extensions is less than the extension of the genus concept.

For example, the daytime and night under the genus concept of “day” are fuzzy concepts, while the night is the opposite negation of daytime. The diagram illustrating the extensional relationship between them is shown below (Figure 5).

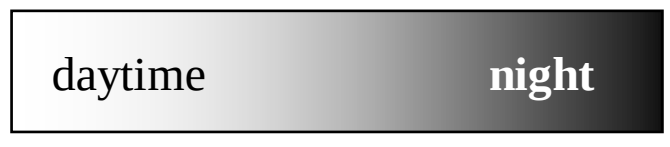

**Fig. 5** The extensional relationship between daytime and night

(6) Intermediary Negation in Fuzzy Concepts (INF)

Characteristics of IFC: Extensions are not clear, opposing sides transition to each other through ‘intermediaries’, and the sum of the extensions equals the extension of the genus concept.

For example, dusk, daytime and night are fuzzy concepts under the genus concept of “day”. Dusk serves as an ‘intermediary’ between daytime and night, and it is the intermediary negation of both daytime and night. The diagram illustrating the extensional relationship between them is shown below (Figure 6).

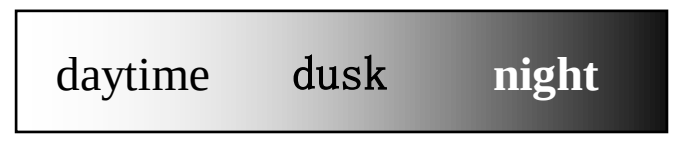

**Fig. 6.** The extensional relationship between daytime, night, and dusk

It is evident that, because concepts are the most fundamental components of knowledge, these three different negations also exist within knowledge itself.

# 3 Logic with three kinds of negation

For the three different types of negation present in the aforementioned knowledge, in order to establish a mathematical foundation (set and logic) that can fully reflect their properties, relations and laws, we proposed a "set SCOI and logic LCOI&PLCOI with contradictory negation, opposite negation and intermediary negation"[33, 34]. In this section, we summarize some basic concepts and main conclusions of LCOI&PLCOI as follows.

## 3.1 LCOI: Propositional Logic with three kinds of negation

The symbols $\neg$, $\urcorner$, and $\sim$ denote "contradictory negation", "opposite negation" and "intermediary negation" respectively. The symbols $\vee$, $\wedge$ and $\to$ denote 'disjunction', 'conjunction' and 'implication', respectively. The symbol "$\vdash$" denote formal deduction.

**Definition 1**. Let $\Im$ be set of atomic proposition. $\forall A\in\Im$, A is called well-formed formula (or formula). If A, B are formulas, then $\neg A$, $\urcorner A$, $\sim A$, $A\to B$, $A\vee B$ and $A\wedge B$ are formulas.

(I) The following formulas as axioms:

(a1) $A\to(B\to A)$
(a2) $(A\to(A\to B))\to(A\to B)$
(a3) $(A\to B)\to((B\to C)\to(A\to C))$
(a4) $(A\to \neg B)\to(B\to \neg A)$
(a5) $(A\to \urcorner B)\to(B\to \urcorner A)$
(a6) $\neg A\to(A\to B)$
(a7) $((A\to \neg A)\to B)\to((A\to B)\to B)$
(a8) $A\to A\vee B$
(a9) $B\to A\vee B$
(a10) $A\wedge B\to A$
(a11) $A\wedge B\to B$
(a12) $\urcorner A\to \neg A\wedge \neg \sim A$
(a13) $\sim A\to \neg A\wedge \neg \urcorner A$

(II) The deduction rules:

[D1] $A_1, A_2, \ldots, A_n \vdash A_i$ $(1\le i\le n)$
[D2] $A\to B, A\vdash B$

The logic calculus formal system determined above is called "propositional logic with contradictory negation, opposite negation and intermediary negation", for short LCOI.

## 3.2 PLCOI: Predicate logic with three kinds of negation

On the basis of LCOI, adding predicates, individual words, quantifiers $\forall$ and $\exists$, as well as the following axioms and deduction rule, we can constitute a predicate logic PLCOI with contradictory negation, opposite negation and intermediary negation.

(I) Axioms:

($\forall$1) $\forall x A(x)\to A(a)$
($\forall$2) $\forall x(A\to B(x))\to(A\to\forall x B(x))$
($\forall$3) $\forall x(A\vee B(x))\to(A\vee\forall x B(x))$
($\exists$1) $A(a)\to\exists x A(x)$
($\exists$1) $\forall x(A(x)\to B)\to\exists x(A(x)\to B)$
($\urcorner\forall$) $\urcorner\forall x A(x)\vdash\dashv\exists x\urcorner A(x)$
($\urcorner\exists$) $\urcorner\exists x A(x)\vdash\dashv\forall x\urcorner A(x)$

(II) Deduction rule:

[D3] If $\Sigma \vdash A(a)$ ($\Sigma$ is the set of formulas), where the individual constant $a$ does not appear in $\Sigma$, then $\Sigma \vdash \forall x A(x)$.

The logic calculus formal system determined above is called "predicate logic with contradictory negation, opposite negation and intermediary negation", for short PLCOI.

The propositional logic LCOI and predicate logic PLCOI are denoted as LCOI&PLCOI. Regarding the semantics of LCOI&PLCOI, we provide an infinity-valued semantic interpretation. Under this semantic interpretation, it can be proven that LCOI&PLCOI is sound and complete.

**Definition 2** (infinite-valued interpretation). Let $\Im$ be set of all formulas in LCOI&PLCOI, $\lambda\in(0, 1)$. $\forall A\in\Im$, mapping $\partial: \Im \to [0, 1]$ is called a $\lambda$-assignment of $\Im$, consists of the individual domain D and the following assignments for each constant symbol, function symbol and predicate symbol in A:

(1) for each constant symbol, assign an object in $D$ to correspond to it;

(2) for each n-variant function symbol, assign a mapping from $D^n$ to $D$ *to* correspond to it;

(3) for each n-variant predicate symbol, assign a mapping from $D^n$ to [0, 1] to correspond to it, and

[1] If A is an atomic formula, $\partial(A)$ takes only one value from [0, 1];

[2] $\partial(A)+\partial(╕A) = 1$;

[3] $$\partial(\sim A) = \begin{cases} \lambda-\frac{2\lambda-1}{1-\lambda}(\partial(A)-\lambda), & \text{when } \lambda\in[½, 1) \text{ and } \partial(A)\in(\lambda, 1] \quad (a) \\ \lambda-\frac{2\lambda-1}{1-\lambda}\partial(A), & \text{when } \lambda\in[½, 1) \text{ and } \partial(A)\in[0, 1-\lambda) \quad (b) \\ 1-\frac{1-2\lambda}{\lambda}\partial(A)-\lambda, & \text{when } \lambda\in(0, ½] \text{ and } \partial(A)\in[0, \lambda) \quad (c) \\ 1-\frac{1-2\lambda}{\lambda}(\partial(A)+\lambda-1)-\lambda, & \text{when } \lambda\in(0, ½] \text{ and } \partial(A)\in(1-\lambda, 1] \quad (d) \\ \partial(A), & \text{other} \quad (e) \end{cases}$$

[4] $\partial(\neg A) = \max(\partial(╕A), \partial(\sim A))$;

[5] $\partial(A\to B) = \max(1-\partial(A), \partial(B))$;

[6] $\partial(A\vee B) = \max(\partial(A), \partial(B))$; $\partial(A\wedge B) = \min(\partial(A), \partial(B))$;

[7] $\partial(\forall xP(x)) = \min_{x\in D}\{\partial(P(x))\}$; $\partial(\exists xP(x)) = \max_{x\in D}\{\partial(P(x))\}$.

**Definition 3** ($\lambda$-tautology). Let $\Gamma$ be set of $\lambda$-assignment of $\Im$, $\forall A\in\Im$. For any $\lambda$-assignment $\partial \in \Gamma$, if $\partial(A) = 1$ then A is called a tautology, if $\partial(A) \geq \lambda$ ($\lambda > ½$), A is called a $\lambda$-tautology, and denoted $\models A$. If there exist $\lambda$-assignment $\partial \in \Gamma$ and $\partial(A) \geq \lambda$, A is said to be $\lambda$- satisfiable,

Based on the above, the following conclusions can be proved.

**Theorem 1** (*Soundness theorem*). Let $\Phi$ ($\Phi\subseteq\Im$) be a set of formulas in LCOI&PLCOI, and A be a formula in LCOI&PLCOI.

(a) If $\vdash A$, then $\models A$.

(b) If $\Phi\vdash A$, then $\Phi\models A$.

**Theorem 2** (*completeness theorem*).

(a) If $\models A$, then $\vdash A$.

(b) If $\Phi\models A$, then $\Phi\vdash A$.

# 4 Modus Tollens based on three types of negation

Modus Tollens (MT) is a classic form of logical inference that is part of deductive reasoning. Its core lies in deriving the negation of the antecedent through the negation of the consequent, thereby ensuring the logical consistency and reliability of the inference process. In MT, the premises are the proposition "If A, then B" and the "negation of B", leading to the conclusion "negation of A". MT has a direct logical relation with negation; the negation is not only key to MT but also reveals the important mechanism of how conditional inference works in reverse.

In classical logic, which includes only one form of negation, one of the most important tautologies is the law of contraposition (LC):

$$A\rightarrow B \equiv \neg B \rightarrow \neg A \qquad \text{(LC)}$$

where, ¬A and ¬B are respectively negations of A and B.

Due to the differentiation of negation in knowledge into contradictory negation ¬, opposite negation ╕, and intermediary negation ~, we discuss Modus Tollens based on these different forms of negation according to the logic LCOI&PLCOI.

## 4.1 Modus Tollens based on distinct negation types: $MT_C$, $MT_O$, $MT_I$

In the logic LCOI&PLCOI [33, 34], we have proved the conclusions in the following formal Theorem 1.

**Theorem 1**. Let A, B be formulas in LCOI&PLCOI.

$\vdash A\rightarrow A$ (1)

$\vdash (A\rightarrow \neg B)\rightarrow((A\rightarrow B)\rightarrow \neg A)$ (2)

$\vdash (A\rightarrow B)\rightarrow((A\rightarrow ╕B)\rightarrow ╕A)$ (3)

$\vdash (A\rightarrow B)\rightarrow((A\rightarrow {\sim}B)\rightarrow {\sim}A)$ (4)

From this, we can further prove the following formal theorems in LCOI&PLCOI.

**Theorem 2**. $A\rightarrow B, \neg B \vdash \neg A$.

Proof:

(a) $A\rightarrow B$
(b) $\neg B$
(c) $\neg B\rightarrow(A\rightarrow\neg B)$ (a1)
(d) $A\rightarrow\neg B$ (b)(c)(D2)
(e) $(A\rightarrow\neg B)\rightarrow((A\rightarrow B)\rightarrow \neg A)$ (2)
(f) $(A\rightarrow B)\rightarrow \neg A$ (d)(e)(D2)
(g) $\neg A$ (a)(f)(D2)

**Theorem 3**. $A\rightarrow B$, ╕B ├ ╕A.

Proof:

(a) $A\rightarrow B$
(b) ╕B
(c) ╕B→(A→╕B) (a1)
(d) A→╕B (b)(c)(D2)
(e) (A→B)→((A→╕B)→╕A) (3)
(f) (A→╕B)→╕A (a)(e)(D2)
(g) ╕A (d)(f)(D2)

**Theorem 4**. $A\rightarrow B, {\sim}B \vdash {\sim}A$.

Proof:

(a) $A\rightarrow B$
(b) ${\sim}B$
(c) ~B→¬B∧¬╕B (a13)
(d) ¬B∧¬╕B (b)(c)(D2)
(e) ¬B∧¬╕B→¬B (a10)
(f) $\neg B$ (d)(e)(D2)
(g) $\neg B\rightarrow(B\rightarrow {\sim}B)$ (a6)
(h) $B\rightarrow {\sim}B$ (f)(g)(D2)

*(i)* $(A\rightarrow B)\rightarrow((B\rightarrow \sim B)\rightarrow(A\rightarrow\sim B))$ (a3)

*(j)* $(B\rightarrow \sim B)\rightarrow(A\rightarrow \sim B)$ $(a)(i)$(D2)

*(k)* $A\rightarrow \sim B$ $(h)(j)$(D2)

*(l)* $(A\rightarrow B)\rightarrow((A\rightarrow \sim B)\rightarrow \sim A)$ (4)

*(m)* $(A\rightarrow \sim B)\rightarrow \sim A$ $(a)(l)$(D2)

*(n)* $\sim A$ $(k)(m)$(D2)

According to the soundness theorem (Theorem 1 in Section 3.2), the formal inferences relationships between the premises and conclusions reflected by the formal theorems in LCOI&PLCOI hold true in deductive reasoning. Therefore, the above theorems 2, 3 and 4 respectively express the following inferences meanings:

(1) With A implies B and the contradictory negation ¬B of B as premises, one can deduce the conclusion that the contradictory negation ¬A of A.

(2) With A implies B and the opposite negation ╕B of B as premises, one can deduce the conclusion that the opposite negation ╕A of A.

(3) With A implies B and the intermediary negation ~B of B as premises, one can deduce the conclusion that the intermediary negation ~A of A.

From this, we propose three variants of Modus Tollens corresponding to (1), (2) and (3), namely "$MT_C$: Modus Tollens based on contradictory negation", "$MT_O$: Modus Tollens based on opposite negation", and "$MT_I$: Modus Tollens based on intermediary negation".

$MT_C$, $MT_O$ and $MT_I$ are indicated below:

| $MT_C$ | $MT_O$ | $MT_I$ |
|---|---|---|
| premise 1: A→B | premise 1: A→B | premise 1: A→B |
| premise 2: ¬B | premise 2: ╕B | premise 2: ~B |
| conclusion: ¬A | conclusion: ╕A | conclusion: ~A |

In the logic LCOI&PLCOI, the formal representations of $MT_C$, $MT_O$ and $MT_I$ as follows (the symbol ⇒ represents deduction):

(C) $MT_C$: A→B, ¬B ⇒ ¬A

(O) $MT_O$: A→B, ╕B ⇒ ╕A

(I) $MT_I$: A→B, ~B ⇒ ~A

## 4.2 Implications in $MT_C$, $MT_O$ and $MT_I$

The implication function, commonly referred to as implication (usually denoted by→), is one of the most important logical connectives in logical inference. In Modus Tollens, implication is core concepts[39, 40]. Therefore, in $MT_C$, $MT_O$ and $MT_I$, we need to further determine the implication in the premises and its relationship with the three different types of negation.

The implication is a binary operation in logic. For the implication → in the premises of $MT_C$, $MT_O$ and $MT_I$, we define it as follows with reference to [39] and [40].

**Definition 1**. A mapping $T$: $[0, 1]^2 \rightarrow [0, 1]$ is called a *t-norm*, if it satisfies

(T1) Boundary conditions, $T(0, 0) = 0$, $T(1, x) = x$, $T(0, x) = 0$;

(T2) Monotony, if $x \le y$ then $T(x, z) \le T(y, z)$, for all $z\in[0, 1]$;

(T3) Commutative, $T(x, y) = T(y, x)$, for all $x, y\in[0, 1]$;

(T4) Associative, $T(x, T(y, z)) = T(T(x, y), z)$, for all $x, y, z\in[0, 1]$.

**Definition 2**. A mapping $I_R$: $[0, 1]^2 \rightarrow [0, 1]$ is called an implication in $MT_C$, $MT_O$ and $MT_I$, if it satisfies

$$I_R(x, y) = sup\{s \in [0, 1] \mid T(x, s) \le y\}, \ \forall x, y \in [0, 1].$$

According to the above definition, it is easy to prove that $I_R$ has the following properties:

(I1) If $x \le y$, then $I_R(x, z) \le I_R(y, z)$, for all $z \in [0, 1]$.

(I2) If $y \le z$, then $I_R(x, y) \le I_R(x, z)$, for all $z \in [0, 1]$.

(I3) $I_R(0, 0) = 1$, $I_R(1, 1) = 1$, $I_R(1, 0) = 0$;

(I4) $I_R(0, z) = 1$ and $I_R(z, 1) = 1$, for all $z \in [0, 1]$.

(I5) $T(x, I_R(x, y)) \le y$.

(I6) $T(x, z) \le y$ if and only if $z \le I_R(x, y)$, for all $z \in [0, 1]$.

In Modus Tollens, negation is typically expressed using classical logic negation. Since the negations $MT_C$, $MT_O$ and $MT_I$ are contradictory negation ($\neg$), opposite negation (╕), and intermediary negation (~) respectively, we use the symbol $\mathrm{N}$ to represent the negation in $MT_C$, $MT_O$ and $MT_I$. Therefore, $\mathrm{N} \in \{\neg$, ╕, ~}.

Examining the properties of Modus Tollens reveals that a reasoning framework capable of utilizing MT for backward reasoning must satisfy the following conditions [40, 41]:

(i) Contains three connectives: implication, negation, and the conjunction of implication and negation.

(ii) With respect to negation, implication should satisfy contraposition symmetry.

Therefore, the implication $I_R$ and negation $\mathrm{N}$ in $MT_C$, $MT_O$ and $MT_I$ should satisfy (i) and (ii) within the framework of LCOI&PLCOI. The conditions are as follows:

With respect to $I_R$ and $\mathrm{N}$,

$$T(x, I_R(x, y)) \le y = T(\mathrm{N}(y), I_R(x, y)) \le \mathrm{N}(x), \text{ for all } x, y \in [0, 1]. \qquad \text{(SC)}$$

From this, the following proposition can be proved.

**Proposition 1**. Regarding the negation $\mathrm{N}$, $I_R$ satisfies the contraposition symmetry: $I_R(x, y) = I_R(\mathrm{N}(y), \mathrm{N}(x))$.

Proof: Since $I_R(x, y) = sup\{s \in [0, 1] \mid T(x, s) \le y\}$, so $I_R(\mathrm{N}(y), \mathrm{N}(x)) = sup\{s \in [0, 1] \mid T(\mathrm{N}(y), s) \le \mathrm{N}(x)\}$. Let $s = I_R(x, y)$, as a result $I_R(x, y) = sup\{I_R(x, y) \in [0, 1] \mid T(x, I_R(x, y)) \le y\}$, $I_R(\mathrm{N}(y), \mathrm{N}(x)) = sup\{I_R(x, y) \in [0, 1] \mid T(\mathrm{N}(y), I_R(x, y)) \le \mathrm{N}(x)\}$. According to (SC), $I_R(x, y) = sup\{I_R(x, y) \in [0, 1] \mid T(x, I_R(x, y)) \le y\} = sup\{I_R(x, y) \in [0, 1] \mid T(\mathrm{N}(y), I_R(x, y)) \le \mathrm{N}(x)\}$. Therefore, $I_R(x, y) = I_R(\mathrm{N}(y), \mathrm{N}(x))$. □

The definition 2 indicates that implication $I_R$ is unrelated to negation $\mathrm{N}$. Hence, we define an implication "*NR*-implication" related to the negation $\mathrm{N}$ as follows:

**Definition 3**. Let T be a *t-norm*. A mapping $I_{NR}$: $[0, 1]^2 \to [0, 1]$, and

$$I_{NR}(x, y) = sup\{s \in [0, 1] \mid T(\mathrm{N}(y), s) \le \mathrm{N}(x)\}, \ \forall x, y \in [0, 1].$$

$I_{NR}$ is called *NR*-implication.

*NR*-implication $I_{NR}$ has the following properties.

**Proposition 2**. For all $x, y \in [0, 1]$,

(a) $I_{NR}(x, y) = I_R(\mathrm{N}(y), \mathrm{N}(x))$;

(b) $T(\mathrm{N}(y), I_{NR}(x, y)) \le \mathrm{N}(x)$;

(c) $T(\mathrm{N}(y), s) \le \mathrm{N}(x)$ if and only if $s \le I_{NR}(x, y)$.

Proof:

(a). Since $I_R(x, y) = sup\{s \in [0, 1] \mid T(x, s) \le y\}$, so $I_R(\mathrm{N}(y), \mathrm{N}(x)) = sup\{s \in [0, 1] \mid T(\mathrm{N}(y), s) \le \mathrm{N}(x)\}$, that is $I_{NR}(x, y)$ by the definition 3.

(b). Suppose $T(\mathrm{N}(y), I_{NR}(x, y)) > \mathrm{N}(x)$. Then $\mathrm{N}(y) > \mathrm{N}(x)$ and $I_{NR}(x, y) > \mathrm{N}(x)$. According to $I_{NR}(x, y) = sup\{s \in [0,1] \mid T(\mathrm{N}(y), s) \le \mathrm{N}(x)\}$, $T(\mathrm{N}(y), s) \le \mathrm{N}(x)$, thus $\mathrm{N}(y) \le \mathrm{N}(x)$. So, contradiction. Therefore, $I_{NR}(x, y) = I_R(\mathrm{N}(y), \mathrm{N}(x))$.

(c). If $T(\mathrm{N}(y), s) \le \mathrm{N}(x)$, then $s = \{s \in [0,1] \mid T(\mathrm{N}(y), s) \le \mathrm{N}(x)\} \le sup\{s \in [0,1] \mid T(\mathrm{N}(y), s) \le \mathrm{N}(x)\} = I_{NR}(x, y)$. If $s \le I_{NR}(x, y)$, then $s \le sup\{s \in [0,1] \mid T(\mathrm{N}(y), s) \le \mathrm{N}(x)\}$. From this, $T(\mathrm{N}(y), s) \le \mathrm{N}(x)$. □

## 4.3 Algorithms of $MT_C$, $MT_O$ and $MT_I$ and their reducibility

From the nature of logical reasoning, inference is a logical process of deriving conclusions from premises.

However, the algorithm for reasoning calculates conclusions through prescribed methods rather than deriving them. The algorithmization of inference can employ different computational models [42, 43].

Based on the reasoning characteristic of Modus Tollens, which derives the conclusion $\neg$A from the premises A $\rightarrow$ B and $\neg$B, the key to establishing an algorithm for MT lies in determining what kind of relational structure to use for the premises A $\rightarrow$ B and $\neg$B. There are two main ideas to this:

(1) Consider $\neg$A as the "a conclusion deduced from the premises A$\rightarrow$ B and $\neg$B". The inference relation is "if A$\rightarrow$ B and $\neg$B, then $\neg$A", and the formal relation is $\neg B \wedge (A \rightarrow B) \rightarrow \neg A$.

(2) According to the law of contraposition, consider $\neg$A as "a conclusion deduced from $\neg$B using A$\rightarrow$ B as a premise". The inference relation is "if A$\rightarrow$ B, then $\neg B \rightarrow \neg A$", and the formal relation is $(A \rightarrow B) \rightarrow (\neg B \rightarrow \neg A)$.

For these two different algorithms, from the perspective of logical syntax, $\neg B \wedge (A \rightarrow B) \rightarrow \neg A$ and $(A \rightarrow B)\rightarrow (\neg B \rightarrow \neg A)$ are different. From the perspective of logical semantics, the former aligns more closely with the inference meaning of MT (Modus Tollens) than the latter.

Regarding the algorithms for $MT_C$, $MT_O$ and $MT_I$, we employ to the first idea and use an algorithm for solving truth values.

Based on the definition of the $\lambda$-assignment $\partial$ in the logic LCOI&PLCOI (Definition 2 in Section 3.2) and the conditions (SC) that $I_R$ should fulfill, we propose the following algorithms for $MT_C$, $MT_O$ and $MT_I$.

(I) Algorithm of $MT_C$

Formal expression： $\neg A = \neg B \wedge (A \rightarrow B)$

Algorithm: $\partial(\neg A) = T(\partial(\neg B), I_R(\partial(A), \partial(B))), \forall \partial(A), \partial(B) \in [0, 1]$.

(II) Algorithm of $MT_O$

Formal expression： ╕A = ╕B $\wedge$ (A$\rightarrow$B)

Algorithm: $\partial$(╕A) = $T(\partial$(╕B), $I_R(\partial(A), \partial(B)))$, $\forall \partial(A), \partial(B) \in [0, 1]$.

(III) Algorithm of $MT_I$

Formal expression： $\sim A = \sim B \wedge (A \rightarrow B)$

Algorithm: $\partial(\sim A) = T(\partial(\sim B), I_R(\partial(A), \partial(B))), \forall \partial(A), \partial(B) \in [0, 1]$.

Here $\partial(\neg A)$, $\partial$(╕A), $\partial(\sim A)$, $\partial(\neg B)$, $\partial$(╕B) and $\partial(\sim B)$ represent the truth values of $\neg$A, ╕A, $\sim$A, $\neg$B, ╕B and $\sim$B, respectively.

In the theory of logical inference, there is no universally accepted standard regarding the merits of reasoning algorithms. However, the "reducibility" of an algorithm is considered the most fundamental requirement of an algorithm [42, 43]. If the negation $\xi$ in MT is an extension of classical negation$*$, then the reducibility of the algorithm concerning MT is generally defined as follows [43]:

**Definition 4**. For an algorithm of MT, if $\xi B = *B$ then $\xi A = *A$, this algorithm is referred to as a reducible algorithm.

According to this definition, we have the following results regarding the reducibility of the algorithms for $MT_C$, $MT_O$ and $MT_I$.

**Theorem 1**. Let $T$ be a *t-norm*, $I_R$ be the implication in $MT_C$. If $I_R \leq I_{NR}$, then the algorithm (I) of $MT_C$ based on the contradictory negation $\neg$ is a reducible algorithm.

Proof: Let $\neg B = *B$. According to MTC's algorithm (I), then

$\partial(\neg A) = T(\partial(\neg B), I_R(\partial(A), \partial(B)))$
$\leq T(\partial(\neg B), I_{NR}(\partial(A), \partial(B))) = T(\partial(*B), I_{NR}(\partial(A), \partial(B)))$
$\leq \partial(*A)$ (by the (b) in the proposition 2)

$\partial(*A) = T(\partial(*B), I_R(\partial(A), \partial(B)))$
$\leq T(\partial(*B), I_{NR}(\partial(A), \partial(B))) = T(\partial(\neg B), I_{NR}(\partial(A), \partial(B)))$
$\leq \partial(\neg A)$ (by the (b) in the proposition 2)

Thus, $\partial(\neg A) = \partial(*A)$, that is $\neg A = *A$. Therefore, regarding the algorithm (I) of $MT_C$, when $\neg B = *B$, it follows that $\neg A = *A$. According to Definition 4, the algorithm of $MT_C$ is a reducible algorithm. □

Similarly, the following theorems can be proved about algorithm (II) for $MT_O$ and algorithm (III) for $MT_I$.

**Theorem 2**. Let $T$ be a *t-norm*, $I_R$ be the implication in $MT_O$. If $I_R \le I_{NR}$, then the algorithm (II) of $MT_O$ based on the opposite negation ╕ is a reducible algorithm.

**Theorem 3**. Let $T$ be a *t-norm*, $I_R$ be the implication in $MT_I$. If $I_R \le I_{NR}$, then the algorithm (III) of $MT_I$ based on the intermediary negation ¬ is a reducible algorithm.

# 5 Counterfactuals and counterfactual reasoning based on three types of negation

Counterfactual and counterfactual reasoning rely on negation. Counterfactuals are constructed based on a hypothetical negation of actual facts, and counterfactual reasoning is a process of reasoning based on counterfactuals. In counterfactuals and counterfactual reasoning, the semantic model of negation is not based on a single form of negation logic. Since negation in knowledge can be distinguished as contradictory negation, opposite negation, and mediating negation, in this section, we discuss counterfactuals and counterfactual reasoning based on these different types of negation within the framework of the logic LCOI&PLCOI.

## 5.1 Counterfactuals based on three types of negation

Counterfactuals are hypothetical statements that are contrary to facts, and a counterfactual is considered relative to a known fact. Counterfactuals are a type of conditional sentence, typically presented in the “if...then...” format. For example:

(1) Fact: It rained yesterday and the ground was wet. Counterfactual: “If the ground hadn't been wet yesterday, then it didn't rain yesterday”.

(2) Fact: If the butter is heated to 150 degrees Fahrenheit, then the butter will melt. Counterfactual: “If the butter didn't melt, then it wasn't heated to 150 degrees Fahrenheit”.

(3) Fact: I didn't stay up late last night and passed the exam today. Counterfactual: “If I had stayed up late last night, then I would not have passed the exam today”.

(4) Fact: It rained yesterday and I took an umbrella. Counterfactual: “If it hadn't rained yesterday, then I wouldn't have taken the umbrella”.

It is evident that counterfactuals have a dependent relationship with facts. The core of counterfactuals is to construct a hypothetical premise contrary to the known facts by negating them, thereby leading to different conclusions. Each counterfactual conditional sentence implicitly refers to a known fact, and its premise is typically a negation of the fact. However, due to the semantic intricacies of its premises, such as the complexity of possible world semantics, the dependency on causal inference, the implicit nature of background knowledge, non-monotonicity, and so on, such negation does not always equate to logical negation [44,45,46,13].

Logical negation is a fundamental concept in logic, referring to the negation of a proposition or statement. It is a unary operation that acts on a single proposition to produce a new proposition whose truth value is opposite to that of the original proposition [1, 45]. Therefore, we use the presence of logical negation as a criterion to differentiate counterfactual conditionals into the following two types:

- Counterfactual conditionals with logical negation.
- Counterfactual conditionals without logical negation.

The so-called counterfactual conditional with logical negation is relative to a factual conditional sentence. Its premise is the logical negation of the conclusion of the factual conditional sentence, and its conclusion is the logical negation of the premise of the factual conditional sentence. Aside from these, all other counterfactual conditionals are considered counterfactuals without logical negation. Based on the above classification, the counterfactuals in (1) and (2) are counterfactual conditionals with logical negation, while those in (3) and (4) are counterfactual conditionals without logical negation.

In the following discussion on counterfactuals and counterfactual reasoning, we will focus only on counterfactual with logical negation and counterfactual reasoning.

According to the definition of the logical negation [33, 34], the contradictory negation ¬, opposite negation ╕, and intermediary negation ~ in the logic LCOI&PLCOI are three different types of logical negation. Therefore, within the framework of the logic LCOI&PLCOI, we propose the following three types of counterfactual conditionals based on the logical negations ¬, ╕ and ~.

Let a factual conditional be $F$: A→B. The formal representations of the three types of counterfactual conditionals based on logical negations ¬, ╕ and ~ are as follows:

$C_{\neg}$: ¬B → ¬A

$C_{╕}$: ╕B → ╕A

$C_{\sim}$: ~B → ~A

where, $C_{\neg}$ is called the 'counterfactual conditional based on contradictory negation', $C_{╕}$ is the 'counterfactual conditional based on opposite negation', and $C_{\sim}$ is the 'counterfactual conditional with intermediary negation'.

Three types of counterfactual conditionals based on different logical negations are commonly found in reality.

**Example 1**. Fact F1: For a car moving at a constant speed, if fuel is added to the car, it will accelerate. There are following counterfactuals:

(1) If the car did not accelerate, then it was not refueled.

(2) If the car decelerated, then it (had) reduced fuel.

(3) If the car neither accelerated nor decelerated (maintaining the original speed), then it neither refueled nor reduced fuel.

According to the logic LCOI&PLCOI, they can be formally represented as follows:

F1: A→B

A: the car is refueled; ¬A: the car is not refueled; ╕A: the car has been defueled; ~A: the car neither refueled nor defueled.

B: the car accelerates; ¬B: the car does not accelerate; ╕B: the car decelerates; ~B: the car neither accelerates nor decelerates (i.e., maintains its original speed).

Thus, the formal representations of the counterfactual conditionals (1), (2) and (3) are as follows:

(1) ¬B → ¬A

(2) ╕B → ╕A

(3) ~B → ~A

**Example 2**. Fact F2: In the function $y = 2x$, if $y$ is a positive even number, then $x$ is a positive integer. Thus, the following counterfactuals exist:

(4) If $x$ is not a positive integer, then $y$ is not a positive even number.

(5) If $x$ is a negative integer, then $y$ is a negative even number.

(6) If $x$ is neither a positive integer nor a negative integer, then $y$ is neither a positive even number nor a negative even number.

In the same way, they can be formally represented as follows:

F1: A→B

A: $y$ is a positive even number; ¬A: $y$ is not a positive even number; ╕A: $y$ is a negative even number; ~A: $y$ is neither a positive even number nor a negative even number.

B: $x$ is a positive integer; ¬B: $x$ is not a positive integer; ╕B: $x$ is a negative integer; ~B: $x$ is neither a positive integer nor a negative integer.

As in Example 1, the formal representations of the counterfactual conditionals (4), (5) and (6) are as follows:

(4) $\neg B \to \neg A$
(5) $╕B \to ╕A$
(6) $\sim B \to \sim A$

## 5.2 Counterfactual reasoning based on three types of negation and their algorithms

Counterfactual Reasoning (CR) is a process of reasoning based on counterfactuals. According to the three types of counterfactuals based on different logical negations mentioned above, within the framework of the logic LCOI&PLCOI, we propose three types of counterfactual reasoning based on different logical negations: "$CR_{\neg}$: counterfactual reasoning based on contradictory negation $\neg$", "$CR_{╕}$: counterfactual reasoning based on opposite negation ╕", and "$CR_{\sim}$: counterfactual reasoning based on intermediary negation ~".

For these three types of counterfactual reasoning based on different logical negations, we use Example 1 as an illustration. According to the counterfactuals (1), (2) and (3) in Example 1, the three types of counterfactual reasoning based on different logical negations are as follows:

1) For a car moving at a constant speed, if the car is refueled then it accelerates, and the car did not accelerate; therefore, the car was not refueled.
2) For a car moving at a constant speed, if the car is refueled then it accelerates, and the car decelerates; therefore, the car has reduced fuel.
3) For a car moving at a constant speed, if the car is refueled then it accelerates, and the car neither accelerates nor decelerates; therefore the car neither refuels nor reduces fuel.

Here, 1), 2) and 3) are $CR_{\neg}$, $CR_{╕}$ and $CR_{\sim}$, respectively.

They can be formally expressed as follows:

$CR_{\neg}$: $A \to B, \neg B \Rightarrow \neg A$
$CR_{╕}$: $A \to B, ╕B \Rightarrow ╕A$
$CR_{\sim}$: $A \to B, \sim B \Rightarrow \sim A$

It can thus be seen that $CR_{\neg}$, $CR_{╕}$ and $CR_{\sim}$ are respectively the same as the formal expressions (C), (O) and (I) of $MT_C$, $MT_O$ and $MT_I$ in Section 4.1.

Therefore, this indicates an important fact:

- The three types of counterfactual reasoning based on contradictory negation, opposite negation and intermediary negation have the same inference form as the three types of Modus Tollens ($MT_C$, $MT_O$ and $MT_I$) based on contradictory negation, opposite negation and intermediary negation, respectively. In other words, they each have the same inference structure.

The algorithmic implementation of counterfactual reasoning can employ different computational models, including the truth value model [47], causal computational model [48], possible world's model [49], and probabilistic model [50]. For the algorithms of the three counterfactual reasoning ($CR_{\neg}$, $CR_{╕}$ and $CR_{\sim}$) based on different logical negations, we adopt the truth value algorithm.

Since $CR_{\neg}$, $CR_{╕}$ and $CR_{\sim}$ each have the same inference form as $MT_C$, $MT_O$ and $MT_I$ respectively, the algorithms for $MT_C$, $MT_O$ and $MT_I$ can serve as a truth value algorithm for $CR_{\neg}$, $CR_{╕}$ and $CR_{\sim}$. These algorithms aim to solve the truth values $\partial(\neg A)$, $\partial(╕A)$ and $\partial(\sim A)$ of the contradictory negation $\neg A$, opposite negation ╕A and intermediary negation ~A of A in the reasoning conclusions.

According to the algorithms of $MT_C$, $MT_O$ and $MT_I$ ((I), (II) and (III) in Section 4.3) and the definition of the $\lambda$-assignment $\partial$ in the logic LCOI&PLCOI (Definition 2 in Section 3.2), we can solve $\partial(\neg A)$, $\partial(╕A)$ and $\partial(\sim A)$ as follows.

In formal logic, if a proposition is true, then its truth value is defined as 1. Therefore, we can make the following assumption:

If a statement is a fact (i.e., the proposition is true), then its truth value is 1.

Thus, for the counterfactual conditional (1), (2) and (3) in Example 1, since F1 is a fact, the truth value of $A \rightarrow B$ is $\partial(A \rightarrow B) = 1$. Since $I_R$ represents the implication in $MT_C$, $MT_O$ and $MT_I$, so

$$\partial(A \rightarrow B) = I_R(\partial(A), \partial(B)) = 1.$$

According to the $\lambda$-assignment $\partial$, $\partial(A \rightarrow B) = \max(1 - \partial(A), \partial(B)) = 1$, i.e. $I_R(\partial(A), \partial(B)) = \max(1 - \partial(A), \partial(B)) = 1$. From this, according to the algorithms of $MT_C$, $MT_O$ and $MT_I$, then

$\partial(\neg A) = T(\partial(\neg B), I_R(\partial(A), \partial(B))) = \partial(\neg B)$;

$\partial(\urcorner A) = T(\partial(\urcorner B), I_R(\partial(A), \partial(B))) = \partial(\urcorner B)$;

$\partial(\sim A) = T(\partial(\sim B), I_R(\partial(A), \partial(B))) = \partial(\sim B)$.

That is, there is the following truth-value equation:

(i) $\partial(\neg A) = \partial(\neg B)$;

(ii) $\partial(\urcorner A) = \partial(\urcorner B)$;

(iii) $\partial(\sim A) = \partial(\sim B)$.

Therefore, regarding the truth value algorithms of three counterfactual reasoning based on contradictory negation$\neg$, opposite negation $\urcorner$ and intermediary negation$\sim$, the truth value equations (i), (ii) and (iii) indicate that:

- If the reasoning premise $A \rightarrow B$ is true ($\partial(A \rightarrow B) = 1$), the truth values $\partial(\neg A)$, $\partial(\urcorner A)$ and $\partial(\sim A)$ for the reasoning conclusions $\neg A$, $\urcorner A$ and $\sim A$ are identical to the truth values $\partial(\neg B)$, $\partial(\urcorner B)$and $\partial(\sim B)$ for the negations $\neg B$, $\urcorner B$ and $\sim B$ in the premises. This reflects the consistency and accuracy of the truth value algorithms for counterfactual reasoning based on different logical negations.

Below, we take the three counterfactual reasoning from Example 1 as a case to calculate the truth values $\partial(\neg A)$, $\partial(\urcorner A)$ and $\partial(\sim A)$ of the reasoning conclusions $\neg A$, $\urcorner A$ and $\sim A$ according to the definition of the $\lambda$-assignment $\partial$ (Definition 2 in Section 3.2).

In the example 1, since F1 is a fact, so $\partial(A \rightarrow B) = 1$. According to the $\lambda$-assignment $\partial$, $\partial(A \rightarrow B) = \max(1-\partial(A), \partial(B)) = 1$. Because of $\partial(A), \partial(B) \in [0, 1]$, $\partial(A) = 0$ or $\partial(B) = 1$ satisfies $\partial(A \rightarrow B) = 1$.

If $\partial(A) = 0$, then $(\urcorner A) = 1 - \partial(A) = 1$. By the [3] in the definition of $\partial$, there are two situations (b) and (c) for $\partial(\sim A)$ when $\partial(A) = 0$. For situation (b), when $\lambda \in [½, 1)$ and $\partial(A) \in [0, 1-\lambda)$, $\partial(\sim A) = \lambda - \frac{2\lambda - 1}{1-\lambda}\partial(A) = \lambda$. For situation (b), when $\lambda \in (0, ½]$ and $\partial(A) \in [0, \lambda)$, $\partial(\sim A) = 1 - \frac{1-2\lambda}{\lambda}\partial(A) - \lambda = 1-\lambda$. By the [3] in the definition of $\partial$, $\partial(\neg A) = \max(\partial(\urcorner A), \partial(\sim A))$, so $\partial(\neg A) = \partial(\urcorner A) = 1$. Therefore, if $\partial(A) = 0$, then $\partial(\urcorner A) = \partial(\neg A) = 1$, $\partial(\sim A) = \lambda$ ($\lambda \in [½, 1)$), or $\partial(\sim B) = 1 - \lambda (\lambda \in (0, ½])$.

In a similar way, If $\partial(B) = 1$, then $\partial(\urcorner B) = 1 - \partial(B) = 0$. There are two situations (a) and (d) for $\partial(\sim B)$ when $\partial(B) = 1$. For situation (a), when $\lambda \in [½, 1)$ and $\partial(B) \in (\lambda, 1]$, $\partial(\sim B) = \lambda - \frac{2\lambda - 1}{1-\lambda}(\partial(B) - \lambda)$, so $\partial(\sim B) = 1-\lambda$. For situation (d), when $\lambda \in (0, ½]$ and $\partial(B) \in (1-\lambda, 1]$, $(\sim B) = 1 - \frac{1-2\lambda}{\lambda}(\partial(A) + \lambda - 1) - \lambda$. According to the [3] in the definition of $\partial$, $\partial(\neg B) = \max(\partial(\urcorner B), \partial(\sim B))$, so $\partial(\neg B) = \partial(\sim B)$. Therefore, if $\partial(B) = 1$, then $\partial(\urcorner B) = 0$, $\partial(\neg B) = \partial(\sim B) = 1 - \lambda$ ($\lambda \in [½, 1)$), or $\partial(\neg B) = \partial(\sim B) = \lambda (\lambda \in (0, ½])$. By the (i), (ii) and (iii), then $\partial(\urcorner A) = 0$, $\partial(\neg A) = \partial(\sim A) = 1-\lambda$ ($\lambda \in [½, 1)$), or $\partial(\neg A) = \partial(\sim A) = \lambda$ ($\lambda \in (0, ½]$).

In summary, for the three counterfactual reasoning scenarios in Example 1, given the premise that F1 is a fact (i.e., $\partial(A \rightarrow B) = 1$), the truth values $\partial(\neg A)$, $\partial(\urcorner A)$ and $\partial(\sim A)$ of the reasoning conclusions $\neg A$, $\urcorner A$ and $\sim A$:

If $\partial(A) = 0$, then $\partial(\urcorner A) = 1$, $\partial(\neg A) = 1$, $\partial(\sim A) = \lambda$ ($\lambda \in [½, 1)$) or $\partial(\sim A) = 1 - \lambda (\lambda \in (0, ½])$

If $\partial(B) = 1$, then $\partial(\urcorner A) = 0$, $\partial(\neg A) = \partial(\sim A) = 1-\lambda$ ($\lambda \in [½, 1)$) or $\partial(\neg A) = \partial(\sim A) = \lambda$ ($\lambda \in (0, ½]$).

where, $\lambda$ is a variable parameter. The size and variation of $\lambda$ determine the size and range of values for $\partial(\neg A)$, $\partial(\urcorner A)$ and $\partial(\sim A)$, i.e. $\lambda$ is a "threshold" for the range of the values for these truth values. Regarding $\lambda$, its role and significance have been discussed in the set SCOI and the logic LCOI&PLCOI [33, 34].

Similarly, for the three types of counterfactual reasoning in Example 2, compute the truth values ∂(¬A), ∂(╕A) and ∂(~A) of the reasoning conclusions ¬A, ╕A and ~A, as described above.

We believe that the three types of Modus Tollens and counterfactuals and counterfactual reasoning based on contradictory negation, opposite negation and intermediary negation proposed in this study, can be extended to fuzzy Modus Tollens, fuzzy counterfactuals and counterfactual reasoning. We will discuss this in a separate article.

# 6 Conclusions and future work

In Modus Tollens (MT) and the counterfactual and counterfactual reasoning, Negation is an indispensable core concept. The negation in Modus Tollens is typically expressed using the negation from classical logic. In counterfactual and counterfactual reasoning, the semantic model of negation is not based on a single form of negation logic.

Based on the logic LCOI&PLCOI with contradictory negation, opposite negation and intermediary negation, as well as the proofs for the three formal theorems (Theorems 2-4 in Section 4.1), we propose a "MTCOI: Modus Tollens based on the three kinds of negation". The MTCOI consists of three different forms of inference: (1) $MT_C$: MT based on contradictory negation; (2) $MT_O$: MT based on opposite negation; and (3) $MT_I$: MT based on intermediary negation, and discusses the implications in $MT_C$, $MT_O$ and $MT_I$, the algorithms of $MT_C$, $MT_O$ and $MT_I$, and the reducibility of the algorithms.

To introduce the three different types of negation into counterfactuals, we use the presence of logical negation as a criterion to differentiate counterfactual conditionals into two types. We then discuss counterfactuals and counterfactual reasoning that involve logical negation. Since contradictory negation, opposite negation, and intermediary negation in the logic LCOI&PLCOI are three distinct forms of logical negation, we propose three types of counterfactuals and counterfactual reasoning based on contradictory negation, opposite negation, and intermediary negation.

Regarding the relationship between $MT_C$, $MT_O$, $MT_I$ and the three types of counterfactual reasoning based on different logical negations in terms of their inference structure, we argue that the three counterfactuals reasoning based on different logical negations have the same inference form as $MT_C$, $MT_O$ and $MT_I$, respectively. In other words, they share the same inference structure. As a result, the truth value algorithms for $MT_C$, $MT_O$ and $MT_I$ can be as the truth value algorithms for the three counterfactuals reasoning based on different logical negations. The algorithms indicates that if the first premise of the reasoning is true, the truth values of the reasoning conclusions are identical to the truth values of the three negative premises in the reasoning premises, respectively. This reflects the consistency and accuracy of the truth value algorithms.

For future research work, building on this paper, we will explore fuzzy counterfactuals and fuzzy counterfactual reasoning based on the three types of negation, Counterfactual Collaborative Reasoning (CCR), and causal reasoning within knowledge systems, etc.